\documentclass{Interspeech}

\usepackage[utf8]{inputenc}
\usepackage{amsmath}
\usepackage{algorithm}
\usepackage{algorithmic}

\usepackage{graphicx}   
\usepackage{subcaption}   
\usepackage{adjustbox}

\interspeechcameraready

\title{Better Semi-supervised Learning for Multi-domain ASR Through Incremental Retraining and Data Filtering}

\author[affiliation={1}]{Andrés}{Carofilis}
\author[affiliation={1}]{Pradeep}{Rangappa}
\author[affiliation={2}]{Srikanth}{Madikeri}
\author[affiliation={1,3}]{Shashi}{Kumar}
\author[affiliation={1}]{Sergio}{Burdisso}
\author[affiliation={4}]{Jeena}{Prakash}
\author[affiliation={1}]{Esaú}{Villatoro-Tello}
\author[affiliation={1,5}]{Petr}{Motlicek}
\author[affiliation={4}]{Bidisha}{Sharma}
\author[affiliation={4}]{Kadri} {Hacioglu}
\author[affiliation={4}]{Shankar} {Venkatesan}
\author[affiliation={4}]{Saurabh} {Vyas}
\author[affiliation={4}]{Andreas}{Stolcke}

\affiliation{}{Idiap Research Institute , Switzerland, $^3$EPFL, Switzerland, $^4$Uniphore Systems, India \& USA}{}
\affiliation{}{University of Zurich (UZH), Switzerland, $^5$Brno University, Czech Republic}{}

\email{andres.carofilis@idiap.ch}

\begin{document}

\maketitle
\begin{abstract}
Fine-tuning pretrained ASR models for specific domains is challenging when labeled data is scarce. But unlabeled audio and labeled data from related domains are often available. We propose an incremental semi-supervised learning pipeline that first integrates a small in-domain labeled set and an auxiliary dataset from a closely related domain, achieving a relative improvement of 4\% over no auxiliary data. Filtering based on multi-model consensus or named entity recognition (NER) is then applied to select and iteratively refine pseudo-labels, showing slower performance saturation compared to random selection. Evaluated on the multi-domain Wow call center and Fisher English corpora, it outperforms single-step fine-tuning. Consensus-based filtering outperforms other methods, providing up to 22.3\% relative improvement on Wow and 24.8\% on Fisher over single-step fine-tuning with random selection. NER is the second-best filter, providing competitive performance at a lower computational cost.
\end{abstract}

\section{Introduction} 
\label{sec:intro}

Automatic speech recognition (ASR) has achieved significant advances in recent years, largely attributable to the development of end-to-end architectures~\cite{graves2006connectionist_ctc,gulati20_conformer, yao2023zipformer, whisper} and the availability of large volumes of training data. Nonetheless, a recurring obstacle in real-world contexts is the high cost of manually labeling large-scale datasets~\cite{novotney2010cheap}, which limits the ability to expand these models to new domains and scenarios, even though large amounts of unlabelled data are often available~\cite{zhang2022bigssl, bhattacharjee2024minimum}.

In response, semi-supervised learning (SSL) emerges as an alternative to \emph{leverage unlabeled speech data}, generating automatic transcripts or \textit{pseudo-labels} that are then fed back into training~\cite{arazo2020pseudo}. The conventional SSL procedure involves training an initial ``seed'' model on a small supervised corpus, using it to generate pseudo-labels for large amounts of unlabeled speech, and then combining the labeled and pseudo-labeled data to refine the ASR model through a single training step~\cite{DBLP:conf/icassp/ManoharHPK18}. Although this technique has been shown to improve system performance when labeled data are scarce, offering an alternative for domain adaptation~\cite{DBLP:conf/interspeech/Carmantini0R19, DBLP:conf/interspeech/HwangSHS22, DBLP:conf/icassp/MaMKS06,DBLP:journals/taslp/ZhuCWHZY24, srinivasamurthy2017semi}, it also poses risks: unreliable \textit{pseudo-labels} can degrade the performance of the model~\cite{DBLP:journals/taslp/ZhuGCPZY23}.

To mitigate this challenge, {\em incremental} semi-supervised learning gradually incorporates unlabeled data and iteratively regenerates pseudo-labels, so that the model can refine its own predictions over time. Xu et al.\cite{DBLP:conf/interspeech/XuLKHSC20} describe an approach, where a seed model is trained on a small labeled set, then iteratively used to decode segments of unlabeled data and generate new pseudo-labels. These pseudo-labeled segments are reintroduced into the training set to continue with the training process. Other studies extend this idea to multi-genre scenarios where both speaking style and content vary~\cite{khonglah2022incremental}. In contrast, Likhomanenko et al.~\cite{DBLP:conf/interspeech/LikhomanenkoXKS21} propose \emph{slimIPL}, an approach where the ASR model generates pseudo-labels, without an external LM, and dynamically updates them as it learns. These pseudo-labels are stored in a dynamic ``cache” and the model is immediately updated with them. As it improves its recognition capability, it also generates more accurate pseudo-labels for previous samples, replacing obsolete pseudo-labels. In this way, there are no distinctly separate ``pseudo-labeling and re-training” phases, but a unified process. Other recent approaches include momentum pseudo-labeling~\cite{higuchi2022momentum}, which adopts a teacher–student framework with a momentum mechanism to gradually refine pseudo-labels, and continuous SSL from scratch~\cite{DBLP:conf/iclr/BerrebbiCBJL23}, where labeled and unlabeled data are used jointly from the beginning, without a seed model that has been pretrained with labeled data. 

Nonetheless, the success of the incremental schemes heavily depends on the \emph{quality} of the pseudo-labels introduced into the training process. Poor-quality transcripts can reinforce model errors~\cite{DBLP:journals/taslp/ZhuGCPZY23}. To address this, this paper introduces an \emph{incremental SSL pipeline} for multi-domain, data-scarce conditions. Our method incorporates two filtering strategies—to the best of our knowledge, previously unexplored in incremental SSL—that reduce the errors introduced by poor-quality pseudo-labels, and, in addition, leverage auxiliary datasets with related domains to improve the performance of the model in its target domain. Our main contributions are:

\begin{enumerate}
    \item We incorporate a small labeled set from the target domain with an \emph{auxiliary} dataset from a related domain to strengthen the model before unlabeled data is added. In addition, an analysis of the performance impact of each of these two datasets is included.
    \item We propose two filtering methods to identify higher-quality pseudo-labels in incremental SSL: a multi-model consensus approach based on character error rate (CER), and a criterion based on named entity recognition (NER). These strategies aim to select higher-quality pseudo-labels, in contrast with a baseline performing purely random selection. 
    \item We demonstrate the effectiveness of the proposed pipeline with two state-of-the-art end-to-end architectures, a transducer (\emph{Zipformer}~\cite{yao2023zipformer}) and an encoder-decoder model (\emph{Whisper-medium}~\cite{whisper}) on the multi-domain \emph{Wow} corpus\footnote{Website: https://wow-ai.com} (with only 4.5 hours of in-domain labeled data) and on \emph{Fisher English}~\cite{cieri2004fisher} (a corpus of telephone conversations).
\end{enumerate}

The rest of the paper is organized as follows. In Section~\ref{sec:method}, we introduce our pipeline,  Section~\ref{sec:results} describes the experimental setup and results, and we provide our conclusions in Section~\ref{sec:conclusion}.

\section{Incremental Semi-supervised Pipeline}
\label{sec:method}

We propose the pipeline described in Algorithm~\ref{alg:semi-supervised-core}. The process begins with a pretrained base model (\textit{model\_base}) and a manually labeled dataset ($S$). Through an initial fine-tuning step using the dataset $S$, we obtain $\text{model}_0$, which then generates transcriptions (\emph{pseudo-labels}) for the unlabeled set ($U$). These pseudo-labels are then filtered using different strategies (see Section~\ref{sec:filtering}), yielding $K$ subsets ($U_1, U_2, \dots, U_K$) that will be incorporated incrementally. $S$ contains manually labeled datasets, including the recordings of the target domain dataset, called $S\_core$ and, optionally, including auxiliary datasets from domains or acoustic conditions similar enough to be beneficial during the fine-tuning process, which we call $S\_aux$. 

\begin{algorithm}[ht]
\caption{Incremental Semi-supervised Pipeline}
\label{alg:semi-supervised-core}
\begin{algorithmic}[1]

\REQUIRE 
  $model\_base$: pretrained model \\
  $S\_core$: target-domain labeled dataset (Supervised) \\
  $S\_aux$: auxiliary labeled dataset (Supervised, optional) \\
  $U$: unlabeled dataset (Unsupervised) \\

\ENSURE 
  \textit{model\_final}: final incrementally fine-tuned model

\STATE \textbf{Initialization (Iteration 0)} 
\STATE $S \leftarrow S\_core \cup S\_aux$ 
\COMMENT{If $S\_aux$ is not used, $S = S\_core$}
\STATE $ \text{model}_0 \leftarrow \text{Fine-Tune}(\textit{model\_base},\ S)$ 
\vspace{6pt}
\STATE \textbf{Single-pass Pseudo-label Generation \& Filtering}
\STATE $\textit{pseudoData} \leftarrow \text{Decode}(\text{model}_0,\ U)$
\STATE $\{U_1, U_2, \ldots, U_K\} \leftarrow \text{Filter}(\textit{pseudoData})$

\vspace{6pt}
\STATE \textbf{Incremental Iterations:}
\STATE $U'_0 \leftarrow \varnothing$
\FOR{$i = 1$ to $K$} 
  \STATE $U'_i \leftarrow U'_{i-1} \cup U_i$
  \STATE $U'_i \leftarrow \text{Decode}(\text{model}_{i-1},\ U'_i)$
  \STATE $\text{model}_i \leftarrow \text{Fine-Tune}\bigl(\textit{model\_base},\ S\_core \cup U'_i\bigr)$
\ENDFOR

\vspace{6pt}
\STATE $\textit{model\_final} \leftarrow \text{model}_K$
\RETURN $\textit{model\_final}$

\end{algorithmic}
\end{algorithm}

According to our experiments, including $S\_aux$ in Iteration~0 helps improve the performance of the $\text{model}_0$, but we did not observe additional improvements from retaining that auxiliary data in subsequent iterations, therefore, $S\_aux$ (as part of $S$) is only used in Iteration~0. These experiments are explained in Section~\ref{sec:results}. In real-world scenarios, it is common for domains with limited data to be complemented with data from related domains~\cite{DBLP:conf/icassp/HwangMHSGQSSBH22,DBLP:conf/icassp/LagosC24}. During each iteration, a new subset $U_i$ is merged into the accumulated buffer $U'_{i-1}$, forming $U'_i$. The previous model ($\text{model}_{i-1}$) then \emph{decodes} $U'_i$, potentially improving the pseudo-labels if the model has learned useful patterns from previous iterations. Finally, we fine-tune \textit{model\_base} using both $S\_core$ and $U'_i$, producing an updated model ($\text{model}_i$). This procedure is repeated until all $K$ incremental subsets have been integrated, resulting in $\textit{model\_final}$. By always including $S\_core$ in each iteration, the model systematically benefits from all available labeled data from the target domain, while introducing new unlabeled subsets in each iteration.

\subsection{Filtering Strategies}
\label{sec:filtering}

A crucial component of our pipeline is the filtering step, where we select the pseudo-labels to mitigate the impact of noisy transcriptions. Below, we detail the three approaches considered:

\subsubsection{Random Selection}
\label{sec:rand}

This simple baseline picks examples without applying any filtering. It serves as a minimal-cost reference for comparison with proposed filtering methods.

\subsubsection{CER-based Consensus}
\label{sec:consensus}

In the consensus approach—similar to ROVER~\cite{fiscus1997post}—multiple models are used to decode the same audio segments. The assumption is that if different models produce similar transcriptions, the pseudo-labels are more likely to be correct. One way to measure the similarity is via the CER among the predicted sequences. Segments with lower inter-model CER are considered more reliable and are used in the pipeline. Our approach takes into account transcriptions from three public pretrained ASR models: Nemo Parakeet 1.1B\footnote{https://huggingface.co/nvidia/parakeet-rnnt-1.1b}, Whisper-medium\footnote{https://huggingface.co/openai/whisper-medium}, and a Zipformer pretrained\footnote{https://huggingface.co/yfyeung/icefall-asr-gigaspeech-zipformer-2023-10-17} on Gigaspeech~\cite{DBLP:conf/interspeech/ChenCWDZWSPTZJK21}. Once we fine-tune the $model\_base$ to obtain $model_0$, we replace whichever public model (Zipformer or Whisper) matches $model_0$. Hence, the final set is $model_0$ plus two public pretrained models. 

Once each unlabeled segment is transcribed by the three models, pairwise CER are computed and averaged. Concretely, if $T_1$, $T_2$, and $T_3$ are the three transcriptions, let $CER(T_i, T_j)$ denote the CER between $T_i$ and $T_j$. Then the average CER ($\overline{CER}$) is given by:
\[
  \overline{CER}
  =
  \frac{
    CER(T_1, T_2) 
    + CER(T_1, T_3)
    + CER(T_2, T_3)
  }{3}.
\]


All segments whose $\overline{CER}$ is \emph{below} a fixed threshold of 5\% are retained, while the rest are discarded. This filtering step is performed only once. The retained segments are then randomly split into incremental subsets (e.g., $U_1, U_2, \ldots$) for fine-tuning.

\subsubsection{NER-based Filtering}
\label{sec:ner}

This strategy employs an open-source\footnote{https://spacy.io/models/en} transition-based neural network~\cite{honnibal-johnson-2015-improved} to identify entities (e.g., person, location, organization) in each pseudo-labeled segment. Any segment containing at least one recognized entity is retained; segments with no detected entities are discarded. The underlying intuition is that the presence of a named entity may indicate a less random or ``hallucinated” transcription, suggesting that the model accurately captured some verifiable element of speech. Conversely, segments in which no entities are found could be incomplete utterances or sufficiently noisy to prevent reliable entity detection. After filtering, the retained segments are randomly distributed across the incremental subsets (e.g., $U_1, U_2, \ldots$).

\textbf{Validation Against WER.} 
To preliminarily validate these filtering strategies, we analyzed a random 100-hour Fisher subset with manual transcripts. After generating pseudo-labels using a Zipformer pretrained on Gigaspeech, we applied each filtering strategy and computed the word error rate (WER) against the reference transcripts. Table~\ref{tab:subset-fisher-filtering-short} shows that the filtered subsets contain higher-quality pseudo-labels. 

\begin{table}[t]
\centering
\caption{WER on a 100-hour Fisher subset, partitioned by each filtering strategy. Pseudo-labels were generated by a Zipformer pretrained on Gigaspeech and compared to the manual labels of each respective partition.}
\label{tab:subset-fisher-filtering-short}
\begin{adjustbox}{max width=\linewidth}
\begin{tabular}{lrrr}
\toprule
\textbf{Filter} & \textbf{\# Hours} & \textbf{\# Segments} & \textbf{WER (\%)} \\
\midrule
Full subset        & 100.0 & 96k & 22.9 \\
$\overline{CER}<$ 5\%     & 17.7  & 19k & 5.6  \\
$\overline{CER}\ge$ 5\%   & 82.3  & 77k & 27.8 \\
NER (contains entity)     & 27.0  & 16k & 17.0 \\
NER (no entity)   & 73.0  & 80k & 24.8 \\
\bottomrule
\end{tabular}
\end{adjustbox}
\end{table}

\begin{figure*}
	\centering
	\begin{subfigure}{0.50\textwidth}
		\includegraphics[width=\linewidth]{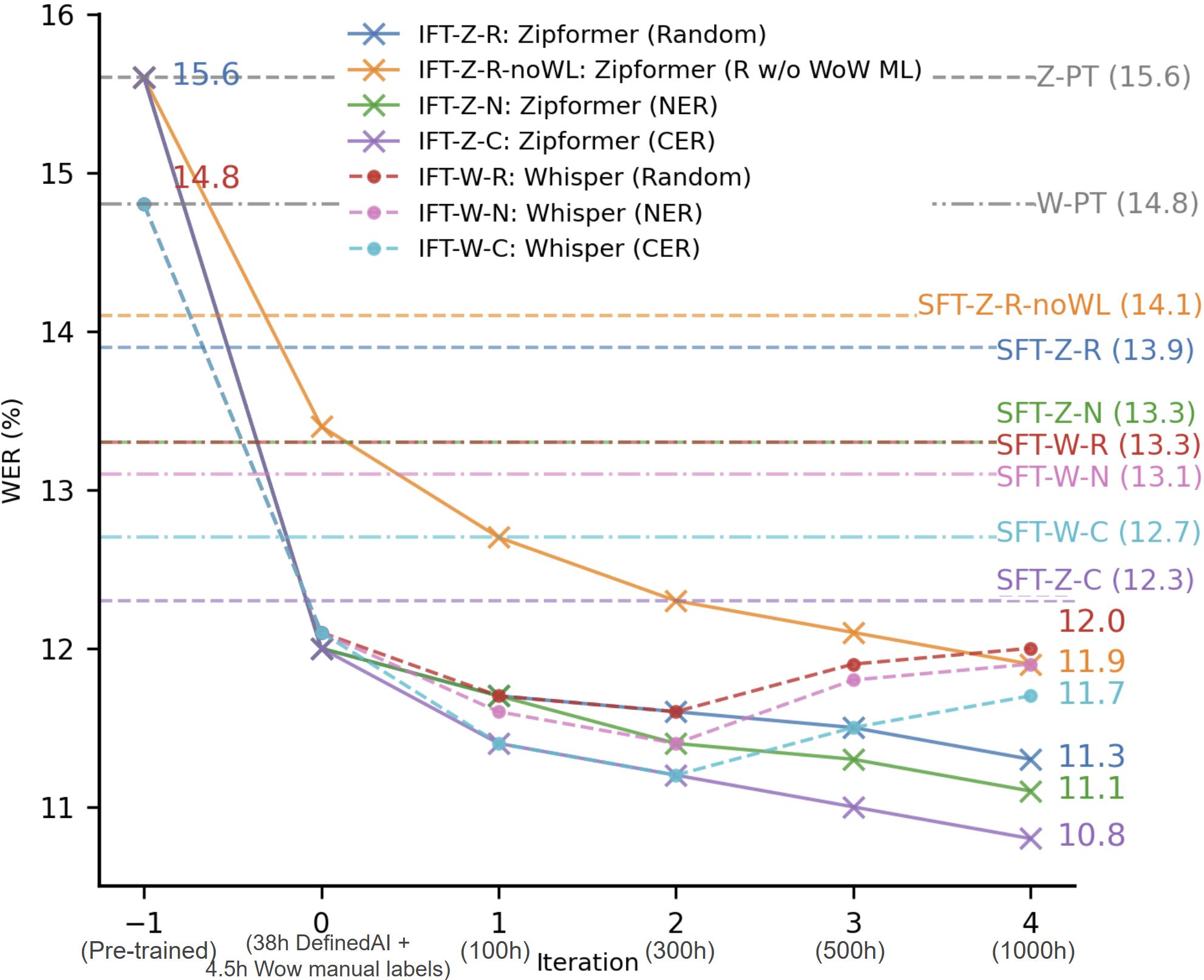}
		\caption{Results on the WoW corpus}
		\label{figure:wow_results}
	\end{subfigure}
	\begin{subfigure}{0.49\textwidth}
		\includegraphics[width=\linewidth]{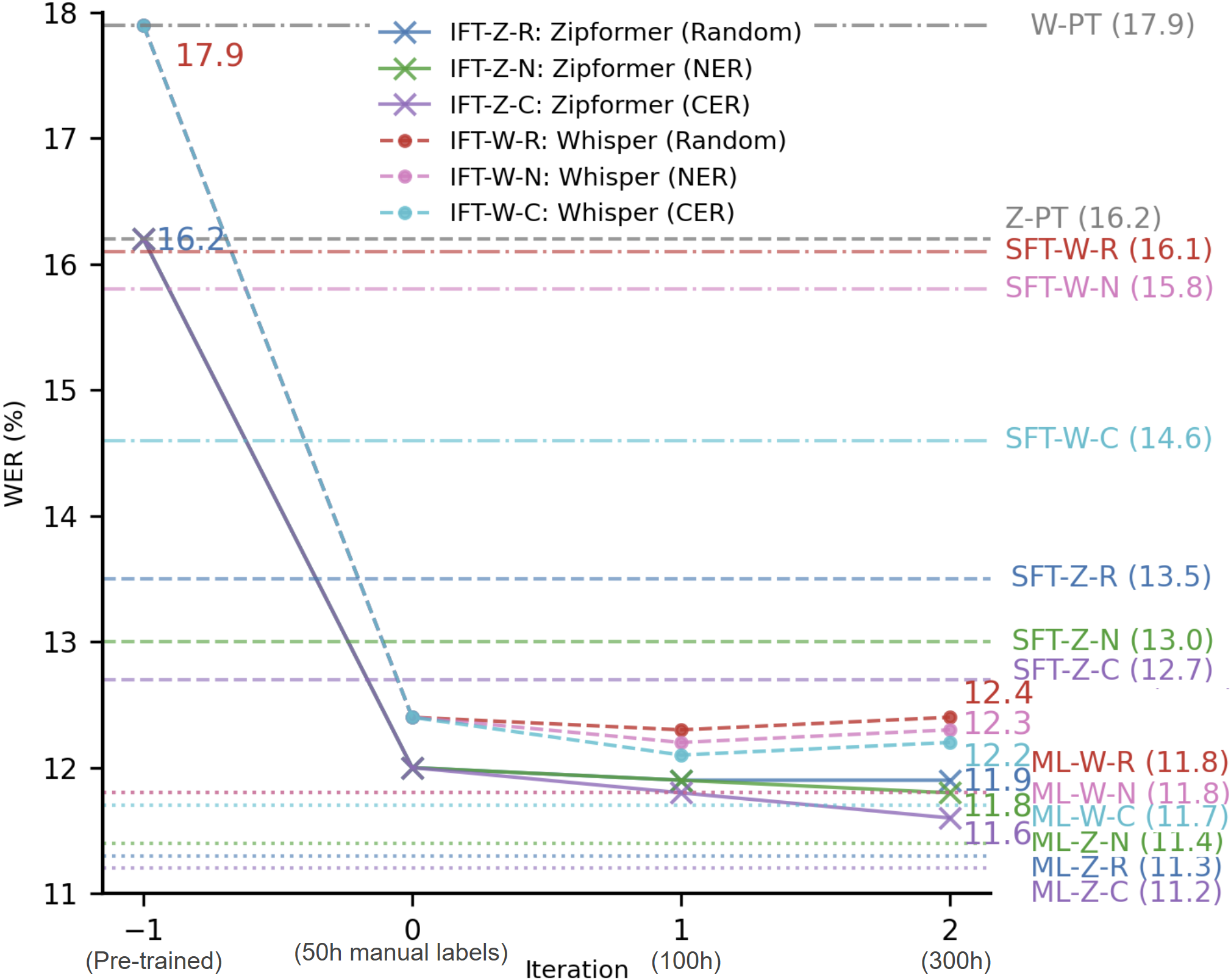}
		\caption{Results on the Fisher English corpus}
		\label{figure:fisher_results}
	\end{subfigure}
	\caption{WER performance on the WoW and Fisher datasets under the proposed incremental pipeline with different pseudo-label filtering strategies. Each result is labeled with an \emph{ID} in the format \{\texttt{Approach}\}-\{\texttt{Model}\}-\{\texttt{Method}\}, where \texttt{Approach} may be IFT (Incremental Fine-Tuning), SFT (Single-step Fine-Tuning), or ML (Manual Labels); \texttt{Model} is Z (Zipformer) or W (Whisper); and \texttt{Method} can be R (Random), N (NER), C (CER), or R-noWL (Random without WoW manual labels). The baselines \texttt{Z-PT} and \texttt{W-PT} are also shown for comparison, corresponding to the Zipformer and Whisper public checkpoints,  where \texttt{PT} refers to ``pretrained". Although the number of hours used is the same, the number of retained segments may vary depending on the filtering strategy.}
	\label{figure:plot_results}
\end{figure*}

\section{Experimental Results}
\label{sec:results}

In this section, we evaluate the performance of our semi-supervised incremental pipeline on two telephony-speech corpora, Fisher English and Wow, using both \emph{Zipformer} and \emph{Whisper-medium}. The results are presented in terms of WER.

\label{subsec:wow_results}

\subsection{Wow Setup}
\label{subsec:wow_setup}

The Wow dataset contains call center conversations across six domains: automotive, auto insurance, medicare, medical, home services, and customer services. Its training set consists of approximately 70,000 conversations, split into 1.35 M segments, for a total of 7,500 hours of audio, but with no ground-truth transcripts. Meanwhile, the test set contains 250 conversations, 26,000 segments, and 18 hours of labeled audio. Additionally, the dataset provides 4.5 hours of labeled speech containing 60 conversations and 5,000 segments. In our experiments, we have considered these 4.5 hours of labeled Wow data as the $S\_core$ subset in the proposed pipeline (see Algorithm~\ref{alg:semi-supervised-core}).

To enhance the performance of $model\_0$ at Iteration~0, we also include an auxiliary subset ($S\_aux$) from the DefinedAI dataset\footnote{Website: https://www.defined.ai}. Specifically, a subset of 38 hours was used. DefinedAI is a call center dataset, with manually annotated labels and covering domains that partially overlap with the domains of Wow—namely, banking, telecom, retail, and insurance. Merging these 38 hours with the 4.5 hours of labeled Wow data produces a total of 42.5 hours of labeled speech ($S$). From the unlabeled Wow training set ($U$), we extracted 995.5 hours of audio for the incremental learning process. This training set was divided into four subsets of 95.5 hours, 200 hours, 200 hours, and 500 hours, added incrementally in Iterations~1--4. Following the results of Khonglah et al.~\cite{khonglah2022incremental}, a saturation of the performance gains can be expected as more unlabeled data is introduced in later iterations. To partially mitigate this saturation, we increased the size of the subsets in some iterations rather than distributing them evenly. Thus, in combination with $S\_core$ the size of the cumulative sets $U'_i$ for Iterations~1--4 is 100 hours, 300 hours, 500 hours, and 1,000 hours, respectively.

We compared the results achieved with our pipeline to a single-step baseline that processes all labeled and unlabeled data—pseudo-labeled by the pretrained models ($model\_base$)—in one single fine-tuning step. For all baselines, the total data volume matches that of the incremental experiments, i.e., the 4.5 hours of Wow labeled data, the 38 hours of DefinedAI, and 995.5 hours of pseudo-labels from the Wow training set were combined, for a total of 1038 hours. In the case of the baseline with random data selection, the same audio segments were used as those used in the incremental learning experiment with random selection. For the NER and CER-based selection baselines, $model\_base$ generates pseudo-labels, which are then filtered using the corresponding strategy, and 995.5 hours are randomly selected.

\subsection{Results on Wow Data}

Figure~\ref{figure:wow_results} shows the WER for Zipformer- and Whisper-medium–based models. As can be seen, CER filtering yields the best performance on Zipformer, where it achieves a 10.8\% WER in Iteration~4—a relative improvement of 4.4\% over Random (11.3\%). NER consistently ranks second (11.1\%), offering smaller but still notable improvement (1.8\%) over Random selection, requiring a lower computational cost than CER (by avoiding multi-model decoding) and remaining competitive. 

For the Whisper model, WER decreases up to Iteration~2 but then increases in Iterations~3 and~4, indicating that Whisper might need more careful hyperparameter tuning or smaller incremental steps. Nevertheless, at Iteration~2, the same behavior can be observed, i.e., CER achieves the best performance, outperforming Random (11.2\% vs. 11.6\%), meanwhile, NER ranks second (11.4\%), suggesting that both filtering methods can produce an improvement across models evaluated. Future work may explore adaptations to the process, to maintain Whisper improvements in subsequent iterations. The incremental approach outperforms the corresponding baselines in all tested filtering methods. Achieving a maximum relative improvement of 18.7\% with Zipformer and random selection.

To measure the effect of retaining the 4.5 hours of Wow ($S\_core$) labeled data at each iteration, an additional experiment (line IFT-Z-R-noWL) was performed in which that subset was removed in the random approach. This increased WER by 5.3\% at Iteration~4 compared to the standard random variant (line IFT-Z-R). This highlights that even a small, high-quality labeled set exerts a marked influence on performance. In addition to the results reported in Figure~\ref{figure:wow_results}, we also evaluated the impact of introducing an auxiliary data set ($S\_aux$) from DefinedAI in Iteration~0, and observed that the WER of Iteration~0 was reduced from 12.5\% (when using only the 4.5 hours of Wow) to 12.0\%. However, in subsequent iterations, no benefit was observed in further use. We observed that in Iteration~1 under Random selection, including DefinedAI led to a WER of 12.3\%, while excluding DefinedAI produced a WER of 11.7\%. A possible explanation is that once enough pseudo-labeled segments have been added, the domain mismatch introduced by DefinedAI may no longer help the model or could increase confusion relative to the newly generated in-domain pseudo-labels.

In addition, a fourth experiment employed the \emph{average log probability} of the tokens of each segment, to choose the higher-pseudo-confidence segments. This metric is equivalent to selecting the lowest-perplexity data, which has shown good results in various natural language processing tasks~\cite{ankner2024perplexed, DBLP:conf/acl/LiuWO22, DBLP:conf/mlnlp/LiuMWLSB22}, where perplexity is defined as $ \mathrm{ppl} = \exp(-\text{avg\_log\_prob}) $. Although it showed a slight improvement over Random selection (13.8\% vs.13.9\% WER in single-step fine-tuning), the improvement was marginal and did not approach the effectiveness of CER or NER. Consequently, it was not included among the proposed filtering methods. Finally, we evaluated whether combining NER and CER filtering on the pseudo-labeled data could yield cumulative gains with Zipformer. In one experiment, a single-step fine-tuning used 50\% of data filtered by each yielded 13.2\%. In another, we performed checkpoint weight averaging between the NER-based (line SFT-Z-N) and CER-based (line SFT-Z-C) baselines, reaching a WER of 12.7\%. Neither approach surpassed the CER-only baseline (12.3\%).

\subsection{Fisher English Setup}
\label{subsec:fisher_setup}

The Fisher English corpus has 1,913 hours of telephone speech; total speech duration is calculated from the Lhotse cutset \cite{zelasko2021lhotse}. Following the Kaldi recipe \cite{povey2011kaldi}, it is split into a training set of 1,906 hours (23,302 recordings, 1,871,669 segments), a dev set of 3 hours (49 recordings, 5,000 segments), and a test set of 3 hours (45 recordings, 5,000 segments). In our experiments, we assume that only 50 hours are labeled while the remaining 1,856 hours are treated as unlabeled data. The 50 hours of labeled data ($S$) also serve as $S\_core$; no $S\_aux$ was used. Meanwhile, a subset of 250 hours is selected from the unlabeled portion ($U$) and divided into 50 hours and 200 hours for Iterations~1 and~2, respectively, following the same size of Wow Setup iterations. The 50 labeled hours and the 250 unlabeled hours of the Random experiments used are the same as those in the Manohar et al.~\cite{DBLP:conf/icassp/ManoharHPK18} paper. In our single-step baseline experiments (lines SFT) the 50 hours of labeled speech are combined with the 250 hours of pseudo-labels in a single fine-tuning step. We also include a \emph{Manual Labels} (lines ML) condition, in which the same 300 hours of audio from Iteration~2 of their corresponding experiments (lines IFT) are used, but using their manual labels.

\subsection{Results on Fisher English}
\label{subsec:fisher_results}

Figure~\ref{figure:fisher_results} shows the WER for the Fisher English experiments. As can be seen, the trends observed with Wow reappear here: the CER approach obtains the best overall performance (11.6\% with Zipformer and 12.2\% with Whisper), followed by NER as the second-best method (11.8\% with Zipformer and 12.3\% with Whisper), and Random ranks last (11.9\% with Zipformer and 12.4\% with Whisper). In all cases, incremental SSL outperforms its single-step fine-tuning baseline, with a maximum improvement of 11.9\% observed when comparing Zipformer with incremental learning and Random selection to its single-step fine-tuning baseline. Notably, the results of the incremental pipelines more closely approximate the results obtained with fully manual labels (lines ML) than those from single-step fine-tuning (lines SFT). For example, Zipformer with CER (11.6\%) exhibits only a 3.6\% relative degradation from the manual-label reference (11.2\%), yet achieves an 8.7\% relative improvement over the single-step baseline (12.7\%). Table~\ref{tab:ift_small_table} presents a comparison between incremental SSL with CER-based filtering versus single-step fine-tuning baseline with random selection. 

\begin{table}[t]
\centering
\caption{Comparison between incremental SSL with CER-based filtering (\textit{IFT-(M)-C}, best iteration) and single-step fine-tuning baseline with random selection (\textit{SFT-(M)-R}). \textit{M} refers to \textit{Model}.}
\label{tab:ift_small_table}
\begin{adjustbox}{max width=\linewidth}
\begin{tabular}{lccccc}
\toprule
\textbf{Dataset} & \textbf{Model} & \textbf{SFT-(\textit{M})-R} & \textbf{IFT-(\textit{M})-C} & \textbf{Relative Impr.\ (\%)} \\
\midrule
WoW    & Zipformer & 13.9 & 10.8 & 22.3 \\
WoW    & Whisper & 13.3 & 11.2 & 15.8 \\
Fisher & Zipformer & 13.5 & 11.6 & 14.1 \\
Fisher & Whisper & 16.1 & 12.1 & 24.8 \\
\bottomrule
\end{tabular}
\end{adjustbox}
\end{table}

\section{Conclusions}
\label{sec:conclusion}

We have introduced an incremental semi-supervised pipeline for ASR, combining a small labeled set with pseudo-labels optionally filtered through inter-model CER-based agreement, or presence of named entities. On Wow and Fisher English data, incremental training outperforms single-step fine-tuning, with CER yielding the best results—surpassing baselines by up to 23.78\% (Wow) and 14.01\% (Fisher). NER is the second-best filter, achieving competitive performance at a lower computational cost. Supplementary data (e.g., DefinedAI) can improve initial models when in-domain labels are scarce. These results show that pseudo-label filtering combined with incremental training significantly improves the performance of ASR models when large amounts of labeled audio are not available. 

\section{Acknowledgments}

This work was supported by the Idiap Research Institute and Uniphore collaboration project.


\bibliographystyle{IEEEtran}
\bibliography{mybib}

\end{document}